\documentclass[sigconf]{acmart}

\usepackage[T1]{fontenc}

\usepackage[utf8]{inputenc}

\usepackage{microtype}

\usepackage{graphicx}

\usepackage{enumitem}
\usepackage{multirow}
\usepackage[ruled,linesnumbered]{algorithm2e}
\usepackage{hyperref}
\usepackage{amsmath}
\usepackage{rotating}
\setenumerate[1]{itemsep=2pt,partopsep=0pt,parsep=\parskip,topsep=5pt}

\newtheorem{definition}{\textbf{Definition}}
\newcommand{\vpara}[1]{\vspace{0.05in}\noindent \textbf{#1 }}
\newcommand{\soay}{\textsc{SoAy}\xspace}
\newcommand{\soaybench}{\textsc{SoAy}Bench\xspace}
\newcommand{\soayllama}{\textsc{SoAy}LLaMA\xspace}
\newcommand{\soaygpt}{\textsc{SoAy}GPT\xspace}
\newcommand{\soayeval}{\textsc{SoAy}Eval\xspace}

\AtBeginDocument{%
  }

\copyrightyear{2025}
\acmYear{2025}
\setcopyright{acmlicensed}
\acmConference[KDD '25]{Proceedings of the 31st ACM SIGKDD Conference on Knowledge Discovery and Data Mining V.1}{August 3--7, 2025}{Toronto, ON, Canada}
\acmBooktitle{Proceedings of the 31st ACM SIGKDD Conference on Knowledge Discovery and Data Mining V.1 (KDD '25), August 3--7, 2025, Toronto, ON, Canada}
\acmDOI{10.1145/3690624.3709412}
\acmISBN{979-8-4007-1245-6/25/08}

\begin{document}

\title{\textsc{SoAy}: A Solution-based LLM API-using Methodology for Academic Information Seeking}


\author{Yuanchun Wang}
\authornote{Work was done when these authors interned at Zhipu AI.}
\orcid{0009-0008-3708-1381}
\affiliation{%
  \institution{Renmin University of China}
  \department{School of Information}
  \institution{Key Laboratory of Data Engineering and Knowledge Engineering, MOE}
  \city{Beijing}
  \country{China}
}
\email{wangyuanchun@ruc.edu.cn}

\author{Jifan Yu}
\authornotemark[1]
\orcid{0000-0003-3430-4048}
\affiliation{%
  \institution{Tsinghua University}
  \department{Institute of Education}
  \city{Beijing}
  \country{China}
}
\email{yujf21@mails.tsinghua.edu.cn}

\author{Zijun Yao}
\authornotemark[1]
\orcid{0000-0002-0288-9283}
\affiliation{%
  \institution{Tsinghua University}
  \department{Department of Computer Science and Technology}
  \city{Beijing}
  \country{China}
}
\email{yaozj20@mails.tsinghua.edu.cn}

\author{Jing Zhang}
\authornote{Corresponding Author.}
\orcid{0000-0003-2019-225X}
\affiliation{%
  \institution{Renmin University of China}
  \department{School of Information}
  \institution{Engineering Research Center of Database and Business Intelligence, MOE}
  \city{Beijing}
  \country{China}
}
\email{zhang-jing@ruc.edu.cn}

\author{Yuyang Xie}
\orcid{0009-0005-4468-3805}
\author{Shangqing Tu}
\authornotemark[1]
\orcid{0009-0008-0640-3413}
\affiliation{%
  \institution{Tsinghua University}
  \department{Department of Computer Science and Technology}
  \city{Beijing}
  \country{China}
}
\email{xieyy21@mails.tsinghua.edu.cn}
\email{tsq22@mails.tsinghua.edu.cn}

\author{Yiyang Fu}
\orcid{0009-0004-2863-1813}
\author{Youhe Feng}
\orcid{0009-0000-7748-3627}
\author{Jinkai Zhang}
\orcid{0009-0005-3053-0082}
\affiliation{%
  \institution{Renmin University of China}
  \department{School of Information}
  \city{Beijing}
  \country{China}
}
\email{fuyiyang2022201505@ruc.edu.cn}
\email{{fyh516, zhangjinkai}@ruc.edu.cn}

\author{Jingyao Zhang}
\orcid{0009-0003-2707-9281}
\author{Bowen Huang}
\orcid{0009-0009-6235-7973}
\author{Yuanyao Li}
\orcid{0009-0002-9977-821X}
\author{Huihui Yuan}
\orcid{0009-0001-1936-6323}
\affiliation{%
  \institution{Zhipu AI}
  \city{Beijing}
  \country{China}
}
\email{{jingyao.zhang, bowen.huang}@aminer.cn}
\email{{yuanyao.li, huihui.yuan}@aminer.cn}

\author{Lei Hou}
\orcid{0000-0002-8907-3526}
\author{Juanzi Li}
\orcid{0000-0002-6244-0664}
\author{Jie Tang}
\orcid{0000-0003-3487-4593}
\affiliation{%
  \institution{Tsinghua University}
  \department{Department of Computer Science and Technology}
  \city{Beijing}
  \country{China}
}
\email{{houlei, lijuanzi, jietang}@tsinghua.edu.cn}


\renewcommand{\shortauthors}{Yuanchun Wang et al.}

\begin{abstract}
Applying large language models (LLMs) to academic API usage shows promise in reducing researchers' efforts to seek academic information. 
However, current LLM methods for using APIs struggle with the complex API coupling commonly encountered in academic queries. 
To address this, we introduce \soay, a solution-based LLM methodology for academic information seeking. 
\soay enables LLMs to generate code for invoking APIs, guided by a pre-constructed API calling sequence referred to as a solution. 
This solution simplifies the model's understanding of complex API relationships, while the generated code enhances reasoning efficiency. 
LLMs are aligned with this solution-oriented, code-based reasoning method by automatically enumerating valid API coupling sequences and transforming them into queries and executable code.

To evaluate \soay, we introduce \soaybench, an evaluation benchmark accompanied by \soayeval, built upon a cloned environment of APIs from AMiner. 
Experimental results demonstrate a 34.58-75.99\% performance improvement compared to state-of-the-art LLM API-based baselines. 
All codes are publicly accessible at \url{https://github.com/RUCKBReasoning/SoAy}.
\end{abstract}



\maketitle

\section{Introduction}
Researchers frequently seek academic information such as a paper's metadata, a scholar's publication record, and the connections between scholars as part of their routine scholarly activities. 
The advent of academic information retrieval systems—including DBLP~\cite{dblp}, CiteSeer~\cite{citeseer}, and Google Scholar~\cite{google-scholar}—and more sophisticated academic mining platforms such as Microsoft Academic Search~\cite{mas} and AMiner~\cite{aminer}, has substantially reduced the complexity of exploring the academic data, which encompasses massive entities like authors, conferences, and papers. 

\begin{figure}[t]
    \centering
    \includegraphics[width=\linewidth]{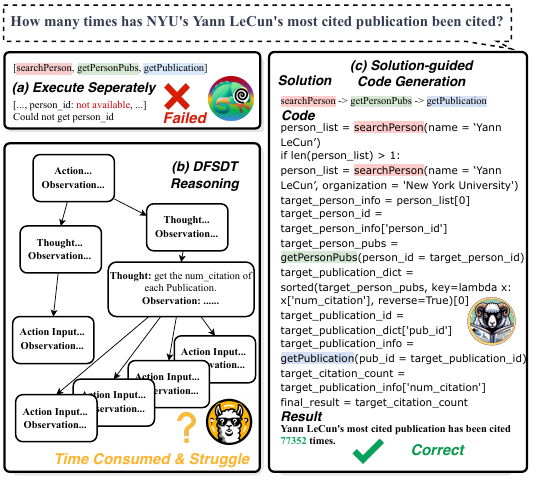}
    \caption{Comparing responses from Chameleon~\cite{chameleon}, ToolLLaMA~\cite{toolllm}, and the proposed \soay, all leveraging AMiner APIs to address the same academic query.}
    \Description{Comparing responses from Chameleon~\cite{chameleon}, ToolLLaMA~\cite{toolllm}, and the proposed \soay, all leveraging AMiner APIs to address the same academic query.}
    \label{fig:intro}
\end{figure}

However, existing systems are constrained by their predefined functionalities and may not fully address the flexible and complex needs of users. 
For example, when a researcher aims to know how many times the representative work of Yann LeCun has been cited, he or she often handcrafts the search keywords into Google Scholar search box, finds and clicks the target item on the result page, and finally seeks the target publication and its citation count.
Large Language Models (LLMs)~\cite{llm-survey}, with their robust reasoning capabilities, offer a promising solution for comprehensively understanding and addressing such nuanced requirements.
One potential approach involves enabling LLMs to leverage academic seeking APIs~\cite{tool-servey}. 
The diverse functionalities offered by these APIs, when combined effectively, hold the potential to address a wide array of user needs. 
However, this adaptation presents two significant challenges:



\noindent\textbf{(1) Coupling: }
One key challenge in adapting large language models (LLMs) to leverage academic retrieval/mining APIs is the inherent parameter coupling within these APIs. For instance, to find the most cited paper of a scholar, the process involves identifying the scholar, retrieving their papers, and then sorting by citation count. This sequence requires the LLM to understand the interdependent calls between the scholar and publication search APIs (i.e., searchPerson and getPersonPubs APIs). Existing methods, such as Chameleon~\cite{chameleon}, as depicted in Figure~\ref{fig:intro}(a), retrieve and execute APIs independently, overlooking the necessary coupling. Consequently, they fail to correctly invoke the getPersonPubs API, which requires a person ID provided by the searchPerson API.

\noindent\textbf{(2) Efficiency: }
Existing methods, such as ToolLLM~\cite{toolllm}, employ a depth-first search-based decision tree (DFSDT) for step-by-step reasoning, which can effectively capture the couplings in the sequence of API calls. However, the numerous steps involved result in multiple inference times, which may be unacceptable to users who are accustomed to the rapid response of existing academic information retrieval systems. For instance, as shown in Figure~\ref{fig:intro}(b), to obtain the most cited paper of a scholar, DFSDT requires one LLM inference per returned paper from the getPersonPubs API to determine its citation count, leading to high inference costs.

\begin{figure*}[ht]
  \centering
  \includegraphics[width=\linewidth]{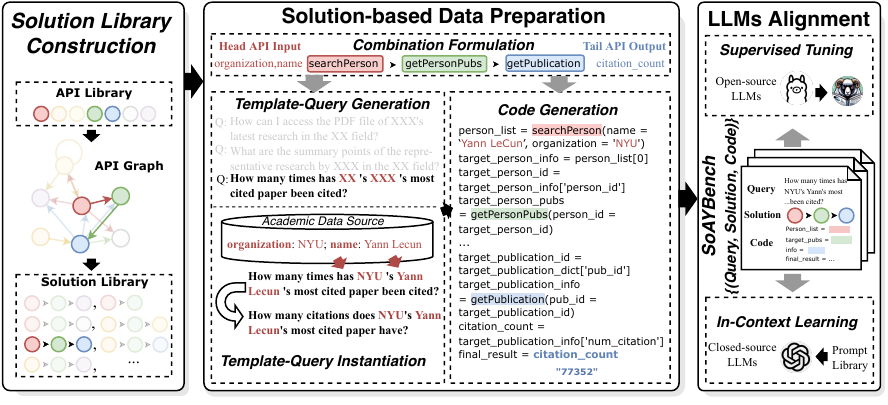}
  \caption{The overall \soay process involves a single academic question case. The main steps of data generation are illustrated on the left and middle sections. Two modes of LLM alignment with the automated triplet dataset are on the right.}
  \Description{The overall \soay process involves a single academic question case. The main steps of data generation are illustrated on the left and middle sections. Two modes of LLM alignment with the automated triplet dataset are on the right.}
  \label{fig:framework}
\end{figure*}

We propose \soay, a \textbf{S\textsc{o}}lution-based \textbf{A}PI-using methodolog\textbf{\textsc{y}} to apply LLMs for academic information seeking. 
Our key innovation is the solution-oriented, code-based method for API calling.
As shown in Figure~\ref{fig:intro}(c), \soay first lets the LLM generate a feasible API calling plan, i.e. solution, based on complex user inputs, and then allows the LLM to generate executable API calling code based on this solution.
The pre-generated solutions help the LLM clarify complex coupling relationships between academic APIs when generating subsequent code. 
By generating the API calling code in a single inference process, \soay achieves efficient inference without repeated reasoning.
To enable LLMs to reliably generate solutions and code in this manner, it is crucial to automatically create a training dataset consisting of (Query, Solution, Code) triplets. 
The process involves using academic APIs published by aminer.org to create a solution library by constructing a dependency graph among the APIs and identifying valid API calling paths. Each solution is then linked with input parameters and output attributes to align more precisely with user intentions. This binding is transformed into a template query and then into Python code. Code correctness is ensured by instantiating the input, executing the code, and verifying the results against expected outputs.


As a result, we automatically develop a dataset containing 3,960 (query, solution, code) triplets, consisting of \soaybench. 
We then introduce \soayeval, an evaluation methodology that takes into account both solution correctness and code execution accuracy.
The experiments demonstrate the effectiveness and efficiency of \soay in utilizing AMiner APIs compared to state-of-the-art LLM API-using baselines, across both in-context learning and tuning modes. 
Furthermore, we validate the versatility of \soay for application in other academic retrieval systems.

The paper contributes: (1) \soay, A novel LLM adaptation method that addresses challenges such as coupling and efficiency in academic information seeking. This method involves automatic data generation, enumerating API calling sequences, transforming them into queries and executable code, and teaching LLMs to use academic APIs. 
(2) \soaybench, a released benchmark dataset, accompanied by the \soayeval evaluation method, to assess LLMs' proficiency in using academic APIs.
(3) Demonstrated effectiveness and efficiency of \soay in the AMiner system, along with its real deployment on an online service.
\section{\soay: Solution-based API-using}
\label{sec:data-preparation}
\subsection{Background}
The process users seek academic information using traditional academic information retrieval systems can be summarized as follow:

\vpara{Intent Decomposition: } To address their complex information needs, users often begin by formulating keywords and entering them into a search interface. For example, to seek the answer of the question illustrated in Figure~\ref{fig:intro}, a user might input ``New York University Yann Lecun'' into the search box of the academic information retrieval system to explore relevant results.

\vpara{Step-by-step Action: } Users then follow a step-by-step process to find the information they need. This involves searching in the search box, browsing through the retrieved scholars, selecting the target scholar, and exploring further by continued browsing and clicking.

\begin{figure*}[t]
  \centering
  \label{fig:solutionlibrary}
  \includegraphics[width=\linewidth]{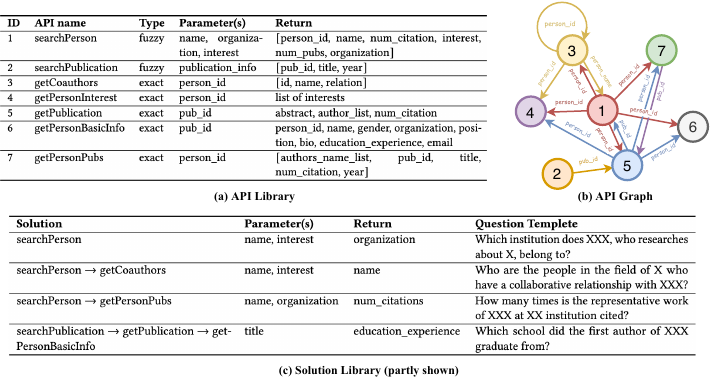}
  \vspace{-0.1in}
  \caption{Illustration of AMiner API Library $L_A$, API Graph $G$ and solution library $L_S$.}
  \Description{Illustration of AMiner API Library $L_A$, API Graph $G$ and solution library $L_S$.}
  \label{fig:aminer-library}
\end{figure*}

This sequence of user actions essentially corresponds to a series of API calls that underpin traditional academic systems. 
For instance, entering a query in the search box translates to calling the searchPerson API as depicted in Figure~\ref{fig:aminer-library}. Clicking on a scholar in the search results corresponds to calling APIs like getPersonPubs to retrieve more detailed scholar information. Finally, clicking on a paper with the most citations to find out its citation count involves calling the getPublication API.
Thus, in this scenario, the API calling sequence is: searchPerson $\rightarrow$ getPersonPubs $\rightarrow$ getPublication. This sequence of API calls is defined as a solution, as detailed in Definition~\ref{eq:api-sequence}.

\subsection{\soay Framework}
To tailor general LLMs to comprehend user requirements and methodically invoke academic APIs correctly and efficiently to access the desired information, we utilize pre-collected solutions $S$ to aid the LLM in crafting Python code that incorporates the necessary API calls. 
To achieve this, abundant training data in the form of (Query, Solution, Code) triplet is essential, facilitating either fine-tuning of existing LLMs or guiding them through in-context learning. 
Illustrated in the left and middle parts of Figure~\ref{fig:framework}, this solution collection and data generation pipeline comprises four key steps:

\vpara{(1) Solution Library Construction (Section~\ref{sec:solution-library}). } Utilizing the provided academic APIs, we construct a graph representing the dependencies among APIs, and traverse this graph to identify valid and concise paths, referred to as solutions. These solutions collectively form a library known as the solution library.

\vpara{(2) Combination Formulation (Section~\ref{sec:combination-formulation}).} To obtain more precise semantics of each solution in the library, we combine all the possible inputs of the initial API and all the outputs of the final API of the solution to formulate a series of combinations.

\vpara{(3) Query Generation (Section~\ref{sec:query-generation}).}
We employ ChatGPT to transform each combination into a template question lacking specific academic entity data. 
These templates are then diversified into various expressions and transformed into specific academic queries, populated with concrete academic entities as input and output from the specialized academic source, the AMiner database.

\vpara{(4) Code Generation (Section~\ref{sec:code-generation}).} 
Finally, we use ChatGPT to generate Python code corresponding to the given combination, which is expected to handle various queries that share the same combination. For each instantiated query, we ensure code correctness by incorporating the query input into the code, executing the code, and verifying the results against the query's ground truth outputs.

Notably, during the process of data creation, we indicate ChatGPT as gpt-3.5-turbo-0613. 
\section{Solution Library Construction}
\label{sec:solution-library}
Given an academic API Library $L_A$, the initial step is to build an API graph where each node represents an API $f$, and a directed edge from node $f_n$ to node $f_m$ signifies a coupling relationship from $f_n$ to  $f_m$ with the intersected attributes $a_{n,m}$.  

\begin{definition} \textbf{API Coupling} defines the frequent dependencies of the inputs and outputs in academic API calling, caused by academic entity correlations' complexity:
\label{eq:api-coupling}
\begin{align}
f_{n} &\xrightarrow{a_{n,m}} f_{m}, \\ if \quad a_{n,m} = i_{m}& \cap o_{n} \quad \text{and} \quad a_{n,m} \neq \emptyset, \nonumber
\end{align}
\noindent where $i_{m}$ denotes the input parameters of $f_m$ and $o_{n}$ denotes the returned attributes of $f_n$. 
\end{definition}
The API coupling $f_n \xrightarrow{a_{n,m}} f_{m}$ means that we can use $f_n$ to output the attribute $a_{n,m}$, and subsequently use $a_{n,m}$ as the input parameter to call $f_m$. 
This is because $a_{n,m}$ is the intersection of the returned attributes $o_n$ of $f_n$ and the input parameters $i_m$ of $f_m$. 
The graph $G$ built from AMiner APIs according to their coupling relationships is shown in Figure~\ref{fig:aminer-library}(b).
By traversing this graph while limiting the maximum number of hops to $H$, we can derive an API calling sequence, referred to as a solution:

\begin{definition} A \textbf{Solution} $S$ is defined as a calling sequence of APIs traversed from the API graph:    
\begin{align}
\label{eq:api-sequence}
   S =  f_{1} \xrightarrow{a_{1,2}} f_{2}  \cdots \xrightarrow{a_{k-1,k}} &f_{k} \cdots \xrightarrow{a_{J-1,J}} f_{J},\nonumber \\ \forall k, f_{k} \in& L_A.
\end{align}
\label{def:solution-defination}
\end{definition}
A solution library $L_S$ denotes a set of valid and simple solutions, ensuring by the following ways.

\vpara{Valid Solution.}
A valid solution can be checked by judging whether the node in the API graph corresponding to the first API in this solution has zero Indegree. 
Note this is specific to academic information seeking systems. The situation may vary for other application systems.

\vpara{Concise Solution.}
For a single query, there may be multiple solutions available. 
The preferred choice is the most concise one, containing the fewest APIs. 
We utilize ChatGPT to verify if two given solutions can obtain the same information. 
The specific prompt used for this verification can be found in our code repo. 
Specifically, each traversed solution is sequentially added to the solution library $L_S$. For each new solution $S$, we check whether a more concise solution already exists in $L_S$ that can achieve the same result as $S$.

Consequently, we've created 16 solutions to establish the solution library $L_S$ for AMiner APIs. 
Figure~\ref{fig:aminer-library}(c) presents a subset of these solutions, including their input parameters, returned attributes, and corresponding question templates.
\section{Soluion-based Data Creation}
For each solution in the solution library, $L_S$, we generate corresponding queries and code. This enables LLMs to effectively perform academic information seeking.

\subsection{Combination Formulation}
\label{sec:combination-formulation}
Typically, a solution may correspond to multiple query intents, each representing a distinct need. Therefore, we formulate combinations derived from a solution to clearly represent each specific intent. We here define a head API $f_1$ as the first API in a solution $S$, while the last API is termed the tail API $f_J$. 
For example, in Figure~\ref{fig:aminer-library}(c), considering the 3-hop solution ``searchPublication $\rightarrow$ getPublication $\rightarrow$ getPersonBasicInfo'', searchPublication is $f_1$ and $f_J$ is getPersonBasicInfo. 
Due to multiple choices of the \textbf{Head API Input $i_1$} and the \textbf{Tail API Output $o_J$}, determining which information (outputs) is sought under what conditions (inputs) remains uncertain even after the solution is established. 
We define a solution-based combination as a specific binding of the head API input, the solution, and the tail API output: $C = (i_1, S, o_J)$, to refine the solution  $S$ into a targeted intent.
For example, given $i_1$ is ``title'' and $o_J$ is ``education\_experience'', the solution  ``searchPublication $\rightarrow$ getPublication $\rightarrow$ getPersonBasicInfo'' can refer to a specific interpretation: querying the alma mater of the first author of the provided publication.

\subsection{Query Generation}
\label{sec:query-generation}
\vpara{Template-Query Generation.}
For each combination \(C\), we define a template question \(Q_T\) to capture its intent. For instance, for the aforementioned combination $C=$(``title'', 
 ``searchPublication $\rightarrow$ getPublication $\rightarrow$ getPersonBasicInfo'', ``education\_experience''), the template question \(Q_T\) could be phrased as, ``Which school did the first author of XXXX graduate from?'' where ``XXXX'' serves as a placeholder for the ``title of a publication'', as shown in Table~\ref{tb:symbols-map}. Additionally, \(Q_T\) could be expressed in variations like, ``What is the alma mater of the first author of XXXX?'' to convey the same intent. To broaden the semantic diversity of an intention, we engage in semantic augmentation by prompting ChatGPT to generate three varied template questions for each combination, enhancing the range of phrasing and nuances.

\vpara{Template-Query Instantiation.}
For each template query \(Q_T\), we substitute the placeholders denoted by uppercase ``X''s with specific attributes such as \textit{person\_id}, \textit{name}, \textit{interest}, \textit{organization}, \textit{bio}, and \textit{educational\_experience} of academic entities from an academic data source to create the instantiated query \(Q\). For instance, using the alma mater query as an example again, the placeholder in \(Q_T\) is replaced with the actual name of a scholar denoted by $i_1'$, resulting in a specific question like ``What is \texttt{scholar name}'s alma mater?''. The actual educational background of this scholar denoted by \(o_j'\), then serves as the ground truth answer to \(Q\).

\begin{table}[t]
\centering
\caption{The semantics of symbols in $Q_T$.}
\small
\begin{tabular}{l|l}
\toprule
Symbol & Semantics \\
\midrule
X & interest \\
XX & organization \\
XXX & name \\
XXXX & publication\_info \\
N & count or numbers \\
Y & Year \\
\bottomrule
\end{tabular}
\label{tb:symbols-map}
\end{table}





\subsection{Code Generation}
\label{sec:code-generation}
So far, a solution $S$ is tailored to a specific intent by associating head API input $i_1$ and tail API output $o_J$ to form a combination $C=(i_1, S, o_J)$, which is linked to three template queries $\{Q_T\}$. Each $Q_T$ is instantiated with real academic entities, resulting in a set of specific queries $\{Q\}$, each with its own input $i_1'$ and corresponding ground truth $o_J'$. 

Building on this data, we then utilize ChatGPT to generate a piece of code corresponding to each combination $C$. This code is designed to retrieve answers for various queries that share the same combination. Using $i_1'$ from a query $Q$ as the input for the code, we execute the code and then compare its output to the ground truth $o_J'$. If the output aligns with the ground truth, the code is considered valid. Any code that fails to produce matching results is excluded from the final dataset.


\section{LLMs Alignment}
To align a general-purpose LLM with the mode of generating corresponding solutions and code based on given problems, we need to prepare relative training data (Section 5.1) and choose the aligning method (Section 5.2).
\subsection{\soaybench--\{(Query, Solution, Code)\}}
\begin{table}[t]
\centering
\caption{Query statistics of the Created Dataset.}
\small
\resizebox{\linewidth}{!}{
\begin{tabular}{c|cccc}
\toprule
Query Type & One-hop & Two-hop & Three-hop & \textbf{Total} \\
\midrule
Scholar & 540 & 1,800 & 540 & 2,980 \\
Publication & 180 & 180 & 720 & 1,080 \\
\textbf{Total} & 720 & 1,980 & 1,230 & \textbf{3,960} \\
\bottomrule
\end{tabular}
}
\label{tb:questionstatistics}
\end{table}
\label{sec:soaybench}
After going through the aforementioned four stages, a substantial amount of high-quality (Query, Solution, Code) data has been automatically created. Table~\ref{tb:questionstatistics} shows the query statistics of the dataset. 
Currently, \soaybench consists of 17 solutions with 44 corresponding combinations. 
Each combination's template query is augmented 3 times and then, through 30 times of query instantiation, results in 90 real queries. 
This process yields a total of $44 \times 3 \times 30 = 3,960$ triplets.
One-fifth of these triplets have undergone expert inspection and adjustment to form a test set, which is published as \soaybench. This test set is supported by a publicly available academic API service, cloned from the online AMiner service. It is used to evaluate the ability of LLMs to search academic information using the AMiner API. The remaining data is used as the training data to support the alignment of LLMs.

\begin{table*}[t]
\centering
\caption{
Overall Performance comparison of the proposed \soay and the baseline methods ToolLLaMA and GTP-DFSDT, as tested on \soaybench. The terms 3.5, 3.5-16K, and 4 are abbreviations for different versions of GPT. 7B and 13B denotes the parameter scale of the LLM. \soaybench is derived from 44 question templates, we calculate each metric within a group of instantiated questions of these templates, and then report the mean and variance across groups with different instantiated entities.
}
\resizebox{\linewidth}{!}{
\begin{tabular}{c|ccccccccc}
\toprule
\multirow{2}{*}{\textbf{Method}} & \multirow{2}{*}{\textbf{Version}} & \multirow{2}{*}{\textbf{Question Type}} & \multicolumn{4}{c}{\textbf{Error Rate}$\downarrow$} & \multirow{2}{*}{\textbf{EM(\%)}} & \multirow{2}{*}{\textbf{ACC(\%)}} & \multirow{2}{*}{\textbf{Score}} \\ 
\cline{4-7}
 & & & \textbf{DS(\%)} & \textbf{WS(\%)} & \textbf{WC(\%)} & \textbf{EE(\%)} & & \\
\cmidrule{1-10}
\multirow{3}{*}{ToolLLaMA} & \multirow{3}{*}{7B} & one-hop & 12.50±8.00 & 24.31±13.26 & 1.39±0.00 & 54.17±16.01 & 7.64±5.20 & 20.14 & \multirow{3}{*}{16.72}   \\
 & & two-hop & 10.10±4.10 & 47.22±12.28 & 0.76±2.27 & 38.13±9.62 & 3.79±2.92 & 13.89 &  \\
 & & three-hop & 11.51±6.53 & 38.10±14.27 & 1.19±3.57 & 43.25±13.07 & 5.95±4.59 & 17.46 & \\
\cmidrule{1-10}
\multirow{9}{*}{GPT-DFSDT} & \multirow{3}{*}{3.5} & one-hop & 55.56±21.06 & 15.28±7.80 & 4.86±0.00 & 21.53±10.67 & 2.78±0.00 & 58.33 & \multirow{3}{*}{43.22} \\
 &  & two-hop & 29.55±11.47 & 34.34±9.23 & 4.29±3.64 & 25.76±8.65 & 6.06±4.11 & 35.61 & \\
 &  & three-hop & 38.10±15.09 & 28.57±11.35 & 3.17±2.50 & 25.00±8.87 & 5.16±6.19 & 43.25 & \\
\cmidrule{2-10}
& \multirow{3}{*}{3.5-16k} & one-hop & 25.69±10.91 & 9.72±5.00 & 2.78±0.00 & 22.92±9.47 & 38.89±15.60 & 64.58 & \multirow{3}{*}{43.67} \\
 &  &two-hop & 16.92±7.76 & 15.91±6.05 & 3.28±1.31 & 46.97±7.13 & 16.92±4.99 & 33.84 & \\
 &  & three-hop & 18.65±7.37 & 15.48±5.63 & 2.78±0.00 & 38.49±10.43 & 24.60±8.53 & 43.25 & \\
\cmidrule{2-10}
 & \multirow{3}{*}{4} & one-hop & 27.78±9.60 & 2.08±0.00 & 4.17±5.00 & 28.47±6.82 & 37.50±10.91 & 65.28 &\multirow{3}{*}{58.16}  \\
 &  & two-hop & 26.26±8.89 & 9.60±4.88 & 17.93±5.40 & 15.15±5.39 & 31.06±9.12 & 57.32 &  \\
 &  & three-hop & 22.22±8.65 & 7.54±4.46 & 17.06±6.96 & 19.05±6.45 & 34.13±9.87 & 56.35 &  \\
\cmidrule{1-10}
\multirow{9}{*}{\soaygpt} & \multirow{3}{*}{3.5} & one-hop & 27.78±8.70 & 15.97±7.73 & 3.47±0.00 & 13.19±7.80 & 39.58±9.12 & 67.36 &\multirow{3}{*}{67.30}  \\
 &  &two-hop & 33.84±4.94 & 9.60±4.75 & 6.06±2.81 & 13.13±7.12 & 37.37±5.06 & 71.21 & \\
 &  & three-hop & 22.22±6.43 & 12.70±5.91 & 9.52±4.42 & 13.10±6.72 & 42.46±6.00 & 64.68 & \\
\cmidrule{2-10}
 & \multirow{3}{*}{3.5-16k} & one-hop & 28.47±11.67 & 15.28±6.12 & 1.39±0.00 & 17.36±7.78 & 37.50±9.07 & 65.97 & \multirow{3}{*}{66.76} \\
 &  & two-hop & 35.86±6.01 & 7.32±3.41 & 5.30±2.18 & 15.91±7.16 & 35.61±4.65 & 71.46 & \\
 &  & three-hop & 23.02±7.16 & 10.32±4.99 & 8.33±3.42 & 17.46±7.37 & 40.87±6.26 & 63.89 & \\
\cmidrule{2-10}
 & \multirow{3}{*}{4} & one-hop & 0.00±0.00 & 0.00±0.00 & 1.39±0.00 & 2.78±0.00 & 95.83±5.70 & 95.83 & \multirow{3}{*}{86.57}  \\
 &  & two-hop & 15.91±4.71 & 1.26±0.00 & 9.34±1.07 & 2.02±1.69 & 71.46±3.74 & 87.37  \\
 &  & three-hop & 6.75±0.00 & 0.40±0.00 & 14.68±1.68 & 1.98±0.00 & 76.19±3.25 & 82.94   \\
 \cmidrule{1-10}
 \multirow{9}{*}{\soayllama} & \multirow{3}{*}{Chat-7B} & one-hop & 0.00±0.00 & 0.00±0.00 & 0.00±0.00 & 0.69±0.00 & 99.31±2.94 & 99.31 &\multirow{3}{*}{85.76}  \\
 &  &two-hop & 0.00±0.00 & 0.00±0.00 & 20.20±3.84 & 2.53±1.97 & 77.27±2.70 & 77.27 & \\
 &  & three-hop & 0.00±0.00 & 0.00±0.00 & 9.92±3.56 & 3.17±2.50 & 86.90±2.72 & 86.90 & \\
\cmidrule{2-10}
 & \multirow{3}{*}{Code-7B} & one-hop & 0.69±0.00 & 0.00±0.00 & 0.69±0.00 & 5.56±4.37 & 93.06±7.50 & 93.75 & \multirow{3}{*}{88.95} \\
 &  & two-hop & 0.25±0.00 & 3.28±0.00 & 7.07±2.75 & 4.80±3.69 & 84.60±5.18 & 84.85 & \\
 &  & three-hop & 0.40±0.00 & 0.00±0.00 & 4.76±2.14 & 5.16±4.57 & 89.68±6.54 & 90.08 & \\
\cmidrule{2-10}
 & \multirow{3}{*}{Code-13B} & one-hop & 0.00±0.00 & 0.00±0.00 & 1.39±0.00 & 0.00±0.00 & 98.61±4.03 & 98.61 & \multirow{3}{*}{\textbf{92.74}}  \\
 &  & two-hop & 0.00±0.00 & 2.27±0.00 & 14.14±2.14 & 0.51±0.00 & 83.08±3.32 & 83.08  \\
 &  & three-hop & 0.00±0.00 & 0.00±0.00 & 2.38±2.86 & 0.40±0.00 & 97.22±4.28 & 97.22   \\
\bottomrule
\end{tabular}
}
\label{tab:overallperformance}
\end{table*}
\label{sec:llm-alignment}
\subsection{Model Alignment}
We provide alignment strategy for both super large closed-source LLMs and tunable open-source LLMs.

\vpara{Closed-source LLMs Alignment.}
\soaygpt is our proposed solution tailored for models like GPT, which, while unable to have their parameters modified, exhibit strong instruction-following capabilities. We engage these closed-source LLMs through in-context learning, leveraging training data from our created dataset to generate prompts. We primarily create three types of prompts: solution, code, and answer generation prompts.

The solution generation prompt is constructed by integrating all solutions explained by their corresponding template queries from the training data. For each solution in the solution library $L_A$, we create a specific code generation prompt by incorporating various instantiated queries and code snippets corresponding to the solution from the training data. Since LLMs may generate solutions not existing in $L_A$, we also create a general code generation prompt for any new solution by including instantiated queries and code snippets of all solutions in $L_A$. The answer generation prompt includes examples of queries, code execution results, and corresponding natural language answers to curate human-friendly responses.

Upon receiving a query, \soaygpt first generates a solution using the solution generation prompt. If the solution can be found in $L_S$, the specific code generation prompt for it is used to generate the code; otherwise, the general code generation prompt is used. The generated code is then executed to produce the result. Finally, the answer generation prompt is used to compose the query and the execution result into a natural language answer. All these prompts are illustrated in Figure~\ref{fig:soaygpt-prompt} in the Appendix.

\vpara{Open-course LLMs Alignment.}
\soayllama, on the other hand, is our alignment strategy for LLaMA~\cite{llama2}, which is an open-source model that we can fine-tune within an acceptable training cost. 
By using the instantiated query of the triplets from the training data as input and integrating the solution and code as output, \soayllama is trained to generate the solution and code in a single inference step based on the given query in a zero-shot setting.
\section{Experiments}
\label{sec:experiments}
In this section, we conduct experiments to validate the effectiveness, efficiency, and practicality of \soay. 
We focus on the following Research Questions: 
(\textbf{RQ1}) Is the \soay method more effective than the baseline method in the task of utilizing the AMiner API to assist academic information retrieval?
(\textbf{RQ2}) Is the \soay method more efficient than the baseline method?
(\textbf{RQ3}) Can the performance of the \soay method in practical applications satisfy users?

We first conduct a detailed comparison of the effectiveness of \soay and previous methods on the \soaybench test set using our carefully designed metric, \soayeval. 
Then, we deploy an application by \soay and validate its effectiveness on collected user queries.

\subsection{Experiments on \soaybench}

\subsubsection{Experimental Settings. }
We propose a detailed and interpretable evaluation method based on \soaybench and conduct comparative experiments with representative baselines.

\vpara{\soayeval.} Enabling large models to use tools for academic information retrieval is typically a multi-step task, with different steps often requiring the model to demonstrate different capabilities. We need detailed evaluation metrics that can help us compare the strengths and weaknesses of different methods and models in specific steps and capabilities.
We propose \soayeval, a detailed evaluation method designed for \soaybench. 
Given a (Query, Solution, Code) triplet from \soaybench, \soayeval first executes the Code in AMiner Environment to get the ground truth answer, then compares the tested result with these ground truth solutions, code, and answer. 
\textbf{(EM)}: Both the solution and answer \textbf{E}xactly \textbf{M}atch the ground truth.
\textbf{(DS)}: The answer is correct, but a \textbf{D}ifferent \textbf{S}olution is generated compared to the ground truth.
\textbf{(WS)}: The answer is wrong due to a \textbf{W}rong \textbf{S}olution.
\textbf{(WC)}: The solution is correct but the answer is wrong, due to a \textbf{W}rong \textbf{C}ode generated for the solution, which can be executed but still leads to a wrong answer.
\textbf{(EE)}: \textbf{E}xecution \textbf{E}rror, which may be caused by the generation of a non-executable code or network error during the request. 
Based on the aforementioned metrics, we define the accuracy metric as follows:
\begin{align*}
    \text{ACC} = \text{EM} + \text{DS},
\end{align*}
\noindent which represents the proportion of correct answers produced by the method under test.
Finally, we assign different weights $w_1$, $w_2$, and $w_3$ to the ACC of one-hop, two-hop, and three-hop of the tested method to compute the overall \textbf{Score}:
\begin{align*}
    \text{Score} = \frac{w_1 \cdot \text{ACC}_1 + w_2 * \cdot \text{ACC}_2 + w_3 \cdot \text{ACC}_3}{w_1 + w_2 + w_3}.
\end{align*}
Notably, we assign $w_1 = 1, w_2 = 2, w_3 = 3$ in our experiment to highlight the multi-hop question-answering ability of the methods under test.

\vpara{Baselines. }
We select GPT-DFSDT, utilizing the same three GPT versions as \soaygpt, and ToolLLaMA-7b~\cite{toolllm} as baselines. 
GPT-DFSDT represents the current state-of-the-art method on multiple general domain tool-utilizing benchmarks\cite{toolllm, stb}. 
ToolLLaMA-7b is an open-source model fine-tuned on a large corpus of tool-utilizing data, achieving comparable results to GPT-DFSDT.

Since there are no benchmark datasets or methods for evaluating the task of academic information retrieval using APIs, we select the two methods mentioned above, which respectively represent the most effective and efficient methods in general domain tool-using tasks. 
We believe these methods can provide relatively strong comparisons for validating the effectiveness of \soay in the domain of academic information seeking.

It is worth noting that results from ToolLLaMA and GPT-DFSDT are in natural language format. 
To ensure a fair comparison using \soayeval, we instruct ChatGPT to extract precise answers from their responses. 
Also, ToolLLaMA and GPT-DFSDT don't have a module to generate a solution or plan before inferring, so we extract the solution from their path of API calls.
For all LLMs, a temperature setting of 0 is used to ensure reproducibility.
\begin{table}[t]
\centering
\small
\caption{Average response time of different methods (second).}
\resizebox{\linewidth}{!}{
\begin{tabular}{c|ccccc}
\toprule
\textbf{Method} & 7B & 13B & 3.5 & 3.5-16k & 4 \\
\midrule
ToolLLaMA & 45.10 & / & / & / & / \\
GPT-DFSDT & / & / & 39.12 & 53.73 & 70.92 \\
\soaygpt & / & / & 6.15 & 6.40 & 26.05 \\
\soayllama-Code & 1.12 & 1.35 & / & / & / \\
\bottomrule
\end{tabular}
}
\label{tb:efficiency}
\end{table}

\subsubsection{Overall Performance \textbf{(RQ1)}. }

Table~\ref{tab:overallperformance} presents the experimental results of \soaygpt, \soayllama, and the other two representative baseline methods on \soaybench. 
\soayllama achieves the highest overall score, with Code-13B Model of 92.74\%, followed by the other \soay method, \soaygpt, which gets 86.57\% with GPT-4 as the backbone.
These results prove the effectiveness of \soay.
With the help of the prepared solution library, \soay describes the complex inter-dependency and diverse execution logic of the LLMs.
This significantly reduces the pressure of LLMs when determining the optimal execution path from the coupling APIs. 

Despite undergoing specialized tool usage training, ToolLLaMA does not score highly. 
This is primarily because ToolLLaMA's training objective is to master the use of general APIs, where the dependencies involved are significantly less complex than those in complicated academic APIs. 
Furthermore, many questions in \soaybench require a solution that incorporates branch or loop logic. 
Facing the need to execute these complicated execution logic, GPT-DFSDT's decision-tree-based inference encounters numerous repetitive sub-nodes, making the solution exceptionally challenging.

\subsubsection{Efficiency \textbf{(RQ2)}. }

We evaluate the efficiency of the proposed method by reporting the average response time of various methods evaluated on \soaybench, as shown in Table~\ref{tb:efficiency}. 
We apply both \soayllama and ToolLLaMA in a network request service format to keep justice with GPT-based methods which should only be accessed through API calls.
Both ToolLLaMA and GPT-DFSDT rely on multiple executions to complete complex API operations, thus consuming a considerable amount of time. 
Implementation through \soay can effectively reduce the inference time of tool calls. 
Even with GPT-4, the average execution time of \soay is nearly one-third of the execution time of GPT-DFSDT with GPT-4. 
Moreover, with a smaller parameter scale, the lightweight \soayllama models show even less execution time, which we consider already meets the requirement for being applied to a real-world service.

\begin{figure}[tb]
  \centering
  \includegraphics[width=0.9\linewidth]{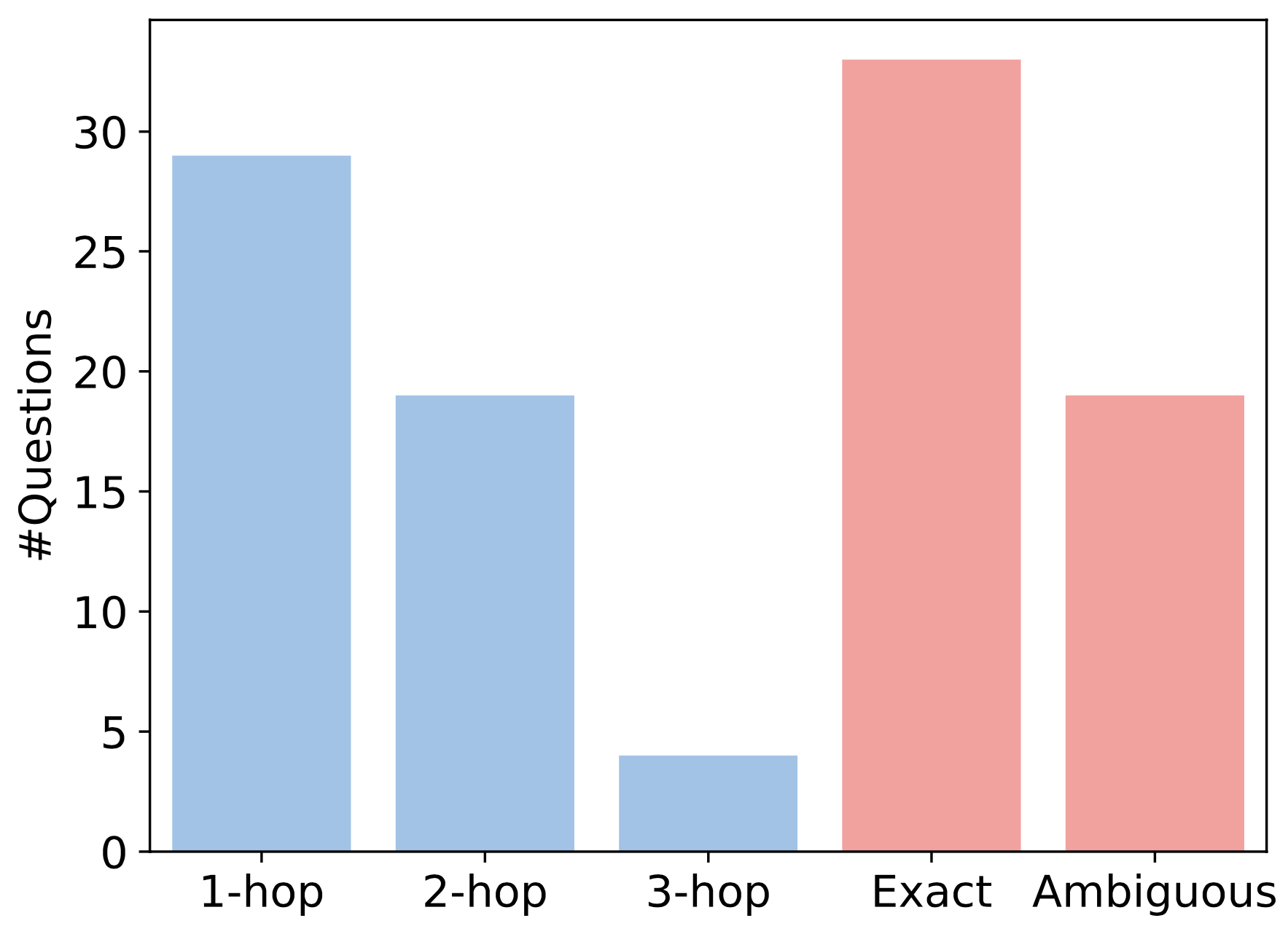}
  \caption{Distribution of questions used for online human evaluation. Questions are categorized by 1, 2, and 3-hop types, further divided into exact and ambiguous categories. The ambiguous category is specific to the online evaluation setting.}
  \Description{Distribution of questions used for online human evaluation. Questions are categorized by 1, 2, and 3-hop types, further divided into exact and ambiguous categories. The ambiguous category is specific to the online evaluation setting.}
  \label{fig:real-questions}
\end{figure}

\subsubsection{Variances. }
Notably, the data fluctuations shown in Table~\ref{tab:overallperformance} do not reflect the variations that would result from repeating the same experiment on identical samples.
Given that the $768$ test samples in \soaybench originate from $44$ template questions, and each template is filled with different entities to create unique question instances, we compute each metric within a group of instantiated questions from the $44$ question templates. 
We then calculate the mean and variance across groups of different instantiated questions. 
Our observations indicate that \soaygpt-4 and \soayllama deliver a more consistently stable performance across various error types, which leverages GPT-4's improved generalization ability and the high-quality training data provided by \soaybench.

\subsubsection{Intriguing phenomena}
Table~\ref{tab:overallperformance} reveals several other intriguing phenomena:
First, while \soay excels with GPT-4, its advantages are less marked on GPT-3.5-turbo and GPT-3.5-16k (hereafter collectively referred to as GPT-3.5 backbones), trailing behind the GPT-DFSDT method with the same GPT-3.5 backbones. Notably, the DS and WS metrics for \soay with GPT-3.5 backbones are quite high (the lower, the better). This can be attributed to GPT-4's superior instruction-following proficiency, which amplifies the impact of \soay's complex prompts on the model adept at instruction following.
Second, \soay methods exhibit superior performance on the two-hop questions than one and three-hop questions, particularly with GPT-3.5 backbones. This is likely because the solution parsing prompts sampled from the \soaybench are biased towards two-hop questions (refer to Table~\ref{tb:questionstatistics} for more information), 
causing the GPT-3.5 backbones to default to two-hop solutions within the \soay framework. On the contrary, GPT-4's robust instruction-following ability enables it to perform consistently across all question types.
Third, GPT-DFSDT exhibits a stronger understanding of one-hop questions when using GPT-3.5 backbones. This is because the general idea of GPT-DFSDT involves breaking down complex questions into simpler one-hop questions, determining the appropriate APIs and their parameters step by step, which enhances its comprehension of one-hop questions.

\subsection{Deployment Evaluation}

Experiments on \soaybench validate the effectiveness and efficiency of \soay. 
However, in practical applications, the questions that users are truly concerned with may differ from those in the dataset. 
To test the user experience of \soay in practical applications, we conduct the following deployment experiment. 

\subsubsection{Settings. }
\label{sec::online-setting}
We deploy \soay as a web application, then gather 52 representative real user queries from the logs, and invite  15 annotators to conduct human evaluation.

\vpara{Queries Detailes. }
The distribution of the 52 gathered queries is illustrated in Figure~\ref{fig:real-questions}. 
We classify them not only into 1, 2, and 3-hop types as \soaybench does but also into either exact or ambiguous questions. 
Ambiguous questions, such as requests for ``comparisons between two scholars'', lack standardized answers and thus yield diverse responses. 

\begin{figure}[t]
  \centering
  \includegraphics[width=\linewidth]{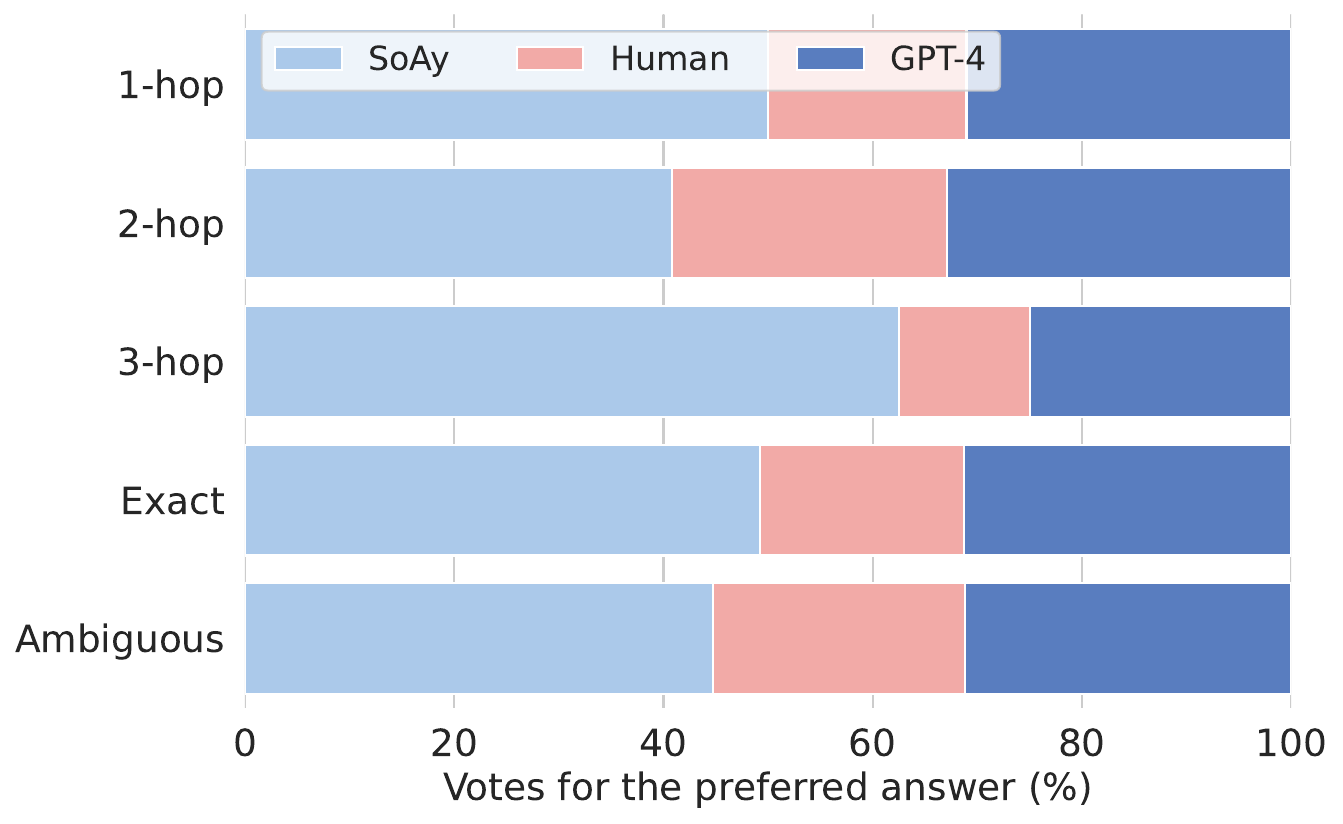}
  \caption{Voting results for authentic online questions. }
  \Description{Voting results for authentic online questions. }
  \label{fig:vote}
\end{figure}

\vpara{Baselines. }
We compare the responses from \soay, GPT-4, and human expert annotated answers on these 52 questions. 
Here GPT-4 indicates GPT-4-1106-preview and \soay indicates \soaygpt with GPT-4 as the backbone.
We do not include GPT-DFSDT, ToolLLaMA, or other tool-using works as comparisons because, based on the results from Section 6.1.3, their efficiency clearly cannot support the requirements of practical deployment.

\begin{figure}[t]
  \centering
  \includegraphics[width=\linewidth]{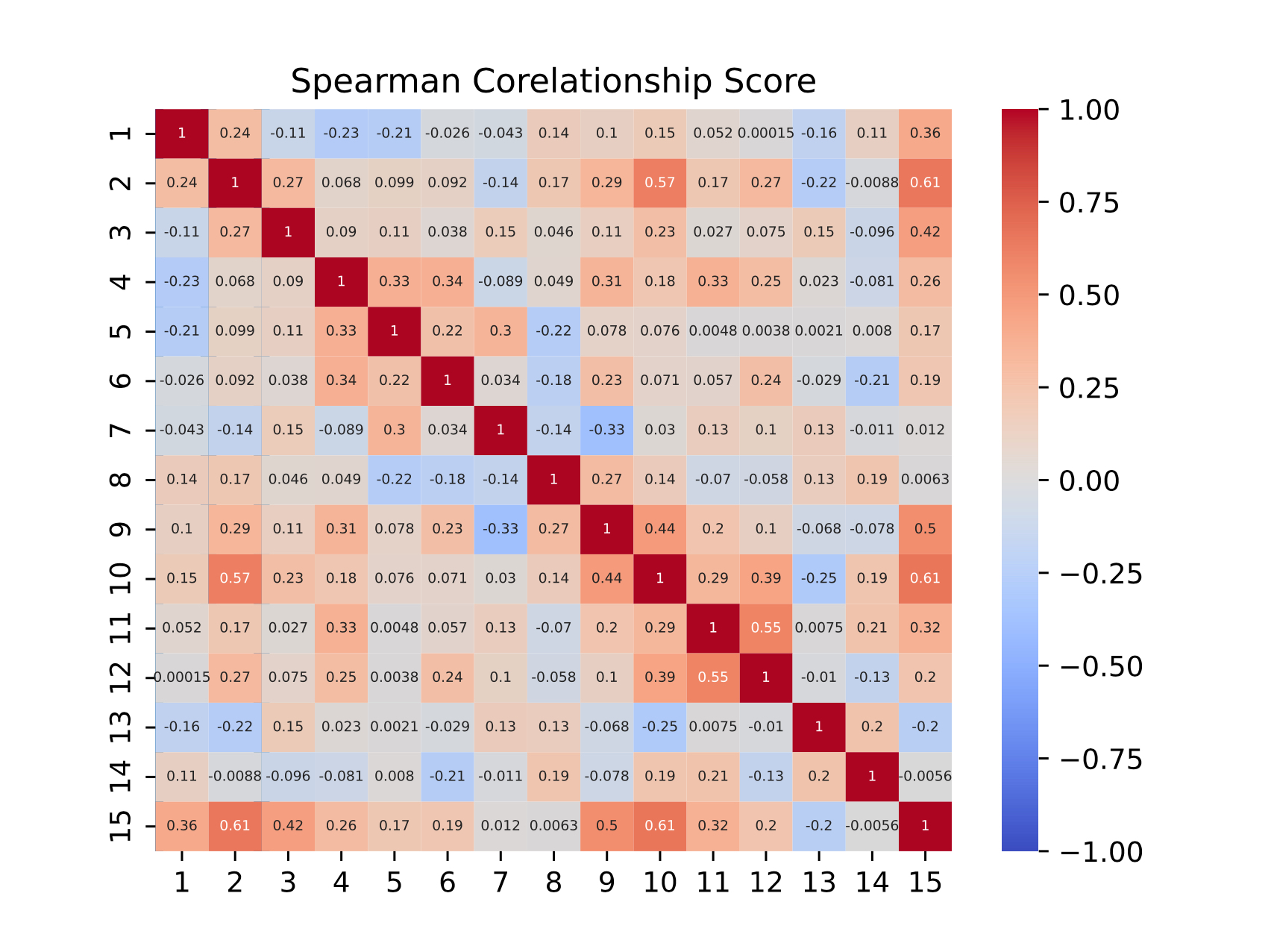}
  \caption{Spearman correlation among 15 annotators}
  \label{fig:spearman}
\end{figure}

\vpara{Metrics. }
We invite 15 annotators to choose their favorite answer from three for each query.
For a balanced and fair assessment of the experimental results, we develop an annotation platform and create a detailed annotation manual. 
This platform allows annotators to shuffle and anonymize the answers from the three methods before each voting session. 
This task did not require the annotators to have expert-level knowledge, so we recruit annotators who are interested in academic tasks through public channels.
As such, the annotators' level of diligence and cognitive abilities might vary significantly, and we aim to filter out annotators who did not annotate seriously, or those whose opinions deviate significantly from the mainstream, through consistency analysis.
We examine the Spearman correlation coefficients among the voting results of the 15 human annotators, as shown in Figure~\ref{fig:spearman}. 
We sum up all the coefficients of each annotator to obtain a trustworthiness score for this annotator. 
We then select the top 10 annotators based on this score and use their votes for further analysis.
For each question, the method selected by the most participants (out of ten) receives one vote. The final vote totals for each method are calculated by summing the votes across all 52 questions.
\subsubsection{Results \textbf{(RQ3)}. }
The summarized results are presented in Figure~\ref{fig:vote}. 
These results show a preference for the \soay method's responses in real-world scenarios across various question types, outperforming both direct GPT-4 and human expert responses.
Notably, \soay's responses are particularly favored for questions with exact answers. 
However, \soay's advantage is less pronounced for ambiguous questions when compared to baseline responses. 
This observation aligns with \soay's design and the assessment objectives of \soaybench, both of which prioritize precise answering through the integration of AMiner API Library resources. Nevertheless, there's a noticeable gap between the nature of real-world inquiries and the templated questions in \soaybench, posing a challenge for \soay in addressing certain scenarios. 
This finding points towards the potential for integrating API calls with model-intrinsic functions for a more robust response system.

\section{Related Works}
\label{sec:related_works}
In the era of Large Language Models (LLMs)~\cite{llm-survey}, the task of academic information retrieval can be accomplished by teaching LLMs to use the APIs of established academic systems~\cite{dblp, citeseer, google-scholar, aminer, mas}. 
Teaching LLMs to handle these APIs is not easy as LLMs' abilities are limited by the extent of knowledge acquired during pre-training~\cite{llm-scaling-law}.
For the LLMs API-using methodology~\cite{tool-servey}, the current technical approaches can be divided into two major directions:

A straightforward idea is to broaden the tool or API-using coverage for LLMs. 
Such efforts have tried to teach LLMs to handle more and more colorful tools/APIs, from the most basic web services~\cite{webgpt, webglm, glm-dialogue, internet-augmented-llm, internet-dialogue, internet-fewshot} to calculators~\cite{toolformer, mathsensei}, KB accessing tools~\cite{lamda2022, gorilla, toolkengpt, toolqa}, code interpreter~\cite{chameleon, pal, pan2023fact, mint}, multi-modal functions~\cite{visual-chatgpt, vipergpt, multi-model-gpt, mm, clova, diffagent} and other scenarios~\cite{let, yao2022webshop, multitool}. 


Some other works have focused on the need to connect multiple tools, requiring LLMs to analyze and plan these relationships~\cite{party, hugginggpt, ART}. 
Examples include Chameleon~\cite{chameleon}, which firstly retrieves relevant APIs based on the given query and then separately executes them to get the responses, ToolChain~\cite{toolchain}, which uses a decision tree for planning, RestGPT~\cite{restgpt}, which combines coarse and fine-grained planning, and ControlLLM~\cite{controlllm}, which reasons based on a graph structure. 
These works often use prompt-based methods with limited manual examples, restricting LLMs' reasoning potential, and can be inefficient due to the high number of reasoning steps needed.
Training-based methods like ToolLLM~\cite{toolllm}, ToolLink~\cite{toolink}, TPTU-v2~\cite{tptuv2}, and OpenAGI~\cite{openagi} can teach LLMs to fully understand some APIs as there's no limitation in the scale of training examples. 
However, these works focus more on APIs that can be individually used like weather query APIs and calculator APIs, but ignore the coupling between the APIs.
Our method, \soay, uses both tuning-free and training-based approaches to solve the coupling problem in academic APIs to seek academic information.


\section{Conclusion}
In this work, we propose an API-using methodology, \soay, to apply LLMs to academic information searching.
\soay generates a solution firstly, which is a sequence of APIs, and then uses this solution to guide the code generation process.
By parsing the solution, \soay can handle the coupling relation between a set of APIs.
Code generation and execution process makes \soay more efficient than previous methods using multi-step reasoning.
The core technique of \soay is the high-quality automatic built dataset, \soaybench, providing 3,960 academic triplets data.
Based on the training set of \soaybench, we build \soayllama and \soaygpt, separately using the tuning method and tuning-free method.
To evaluate the AMiner API-using capability of different models and methods, we propose a comprehensive evaluation strategy, \soayeval.
Experiments show \soay's effectiveness, efficiency, and usability.
\soay has been applied to real-world service and has been accessed more than 54,800 times.
The implementation of \soay to various applications in academic domain highlights the usability of \soay.
In the future, we will continue to enlarge the \soaybench and apply \soay to more scenarios and domains. 
In the future, we will continue to enlarge the \soaybench and enrich the queries in \soaybench based on the real-world application user logs. 
\begin{acks}
This work is supported by the National Key Research \& Develop
Plan (2023YFF0725100) and National Natural Science Foundation of China (62322214, U23A20299, U24B20144, 62172424, 62276270).
This project would like to thank the data annotators at Zhipu AI for their help and support. 
All GPU compute and API costs are provided by Zhipu AI. 
\end{acks}

\clearpage
\bibliographystyle{ACM-Reference-Format}
\balance
\bibliography{bibliography}
\section*{Appendix}
\appendix
\section{Application of \soay}
As of the submission date, \soay has been deployed for 10 months and has been accessed more than 54,800 times in the real-world application at \url{https://www.aminer.cn/} and \url{https://chatglm.cn/main/gdetail/65bf5a99396389a73ace9352}, providing users with accurate academic information and helping them stay informed with up-to-date insights throughout their research process.



\section{\soaygpt}
\label{sec:appendix-soaygpt}
\soaygpt is an alignment method designed for a series of large-scale models that are closed-source yet possess high instruction following capabilities. 
We implement \soaygpt on three distinct versions of GPT, in particular 3.5: gpt-3.5-turbo-0613, 3.5-16k: gpt-3.5-turbo-16k-0613, and 4: gpt-4-0613.
Specifically, \soaygpt employs in-context learning to construct three LLM agents~\cite{agent-survey} (Solution Generation Agent, Code Generation Agent, and Answer Generation Agent) and an environment operable by these agents to complete the alignment process.

Identifying the solution first and then generating the code is less challenging for LLMs, compared to directly code generation, reducing the likelihood of producing incorrect answers. 
Additionally, the advantage of generating the solution and code separately, rather than simultaneously, is that each step allows for more information and examples to be provided within a limited input space.

\begin{figure*}[bht]
  \centering
  \includegraphics[width=0.95\linewidth]{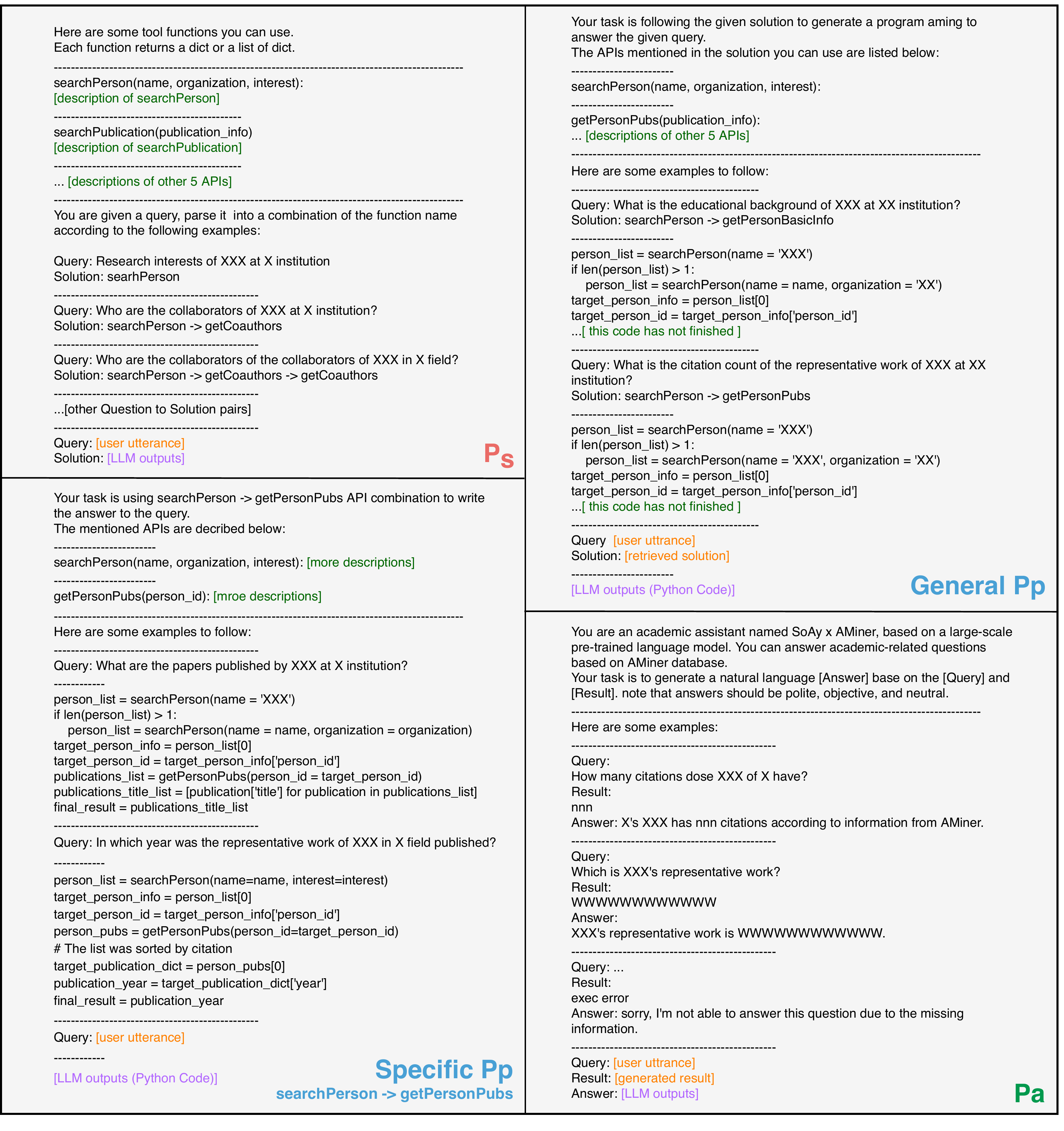}
  \caption{The prompt is designed during the inference of \soaygpt. Specifically:
$P_s$ signifies the prompt for solution generation, which instructs the LLMs to generate a solution from the user's natural language query.
$P_p$ represents the prompt for code generation. This prompt is used to instruct the LLMs to generate code from the user's query and the corresponding solution.
Note that the code generation prompt is tailored for each solution in the solution library $L_S$. If the given solution is found in $L_S$, its specific $P_p$ is used; otherwise, a general $P_p$ is employed. This general prompt is composed by sampling (Query, Solution, Code) triplets from the training data for each solution in $L_S$.
$P_a$ denotes the prompt for answer generation, used to instruct the LLM to generate a natural language answer based on the query and the code execution result.
In the prompt content:
All black text, except the ellipses, represents the original prompt content.
Orange highlighted text includes annotations added for better understanding in the context of creating a prompt.
Purple highlighted text indicates the content awaiting LLM output.}
  \Description{The prompt is designed during the inference of \soaygpt.}
  \label{fig:soaygpt-prompt}
\end{figure*}

The agents and environments are designed as follows.
Prompts for each agent are shown in Figure~\ref{fig:soaygpt-prompt}.

\section{\soayllama}
\label{sec:appendix-soayllama}
\soayllama is designed for supervised fine-tuning of open-source models when training resources are available. During the training stage, \soayllama uses the API description and the query as input for each data sample, and combines the solution and code as the response.
During the inference stage, \soayllama concurrently completes the Planning and Formatting processes to derive solutions and code. 
Similar to \soaygpt, it utilizes the environment to execute and obtain answers based on $K$.
We conduct the fine-tuning on 3 different variants of LLaMA Model, respectively Chat-7B: Llama-2-7b-chat-hf\footnote{\url{https://huggingface.co/meta-llama/Llama-2-7b-chat-hf}}, Code-7B: CodeLlama-7b-Instruct-hf\footnote{\url{https://www.modelscope.cn/models/AI-ModelScope/CodeLlama-7b-hf/summary}}, Code-13B: CodeLlama-13b-Instruct-hf\footnote{\url{https://huggingface.co/meta-llama/CodeLlama-13b-Instruct-hf}}.

\vpara{Dataset. } As mentioned in Section~\ref{sec:llm-alignment}, four-fifth of the instances from \soaybench are utilized here to train the \soayllama.

\vpara{Training Details. }
We train 13B models on 8 NVIDIA A100 GPUs (80GB) for 13 hours to reach a convergent loss, using a batch size of 8. Additionally, we train 7B models on 4 NVIDIA A100 GPUs (80GB) for 7 hours with a batch size of 16.
All the training details are saved in the training log in our Github Repo\footnote{https://anonymous.4open.science/r/SoAy-C025/readme.md}.


\section{\soaybench}
\label{sec:appendix-soaybench}
To assess AMiner API-using capabilities, the foundational AMiner APIs for LLMs to invoke and a test set should be provided. 
However, given the dynamic and specialization nature of academic data, with academic entities-related information rapidly changing, maintaining a test set with static answers proves challenging.
To address this challenge, we clone AMiner's APIs at a specific point in time to create a static service, from which we create a corresponding static test set including 792 meaningful academic questions (one-fifth of the whole \soaybench) and the corresponding ground truth conducted from the cloned APIs.
We extract a subset of the comprehensive AMiner database, which includes over 66,000 entities such as persons and publications, at a specific point in time, September 23, 2023. This subset serves as our experimental evaluation environment, and we clone an API Service based on it. 
Consequently, the results returned by these APIs, given the same parameters, will remain consistent over time. This dataset and the API service provide a foundation for our subsequent construction of the test set.

\section{Implementation on Other Scenarios}
Although \soay is developed for the AMiner API, it can be seen as a general method for enabling large language models (LLMs) to use domain-specific APIs.
To verify this generality, and also to prove that similar coupling phenomena, like those observed with the AMiner API, also exist in other domain-specific APIs, we deploy \soay in two other scenarios that provide information-seeking APIs, OpenLibrary and CrossRef.
Due to GPU resource limitations, we did not train separate models for these two new scenarios. 
Instead, we implemented corresponding \soaygpt models to validate the feasibility of setting up \soay in a new API domain. 
Details can be found at \url{https://github.com/RUCKBReasoning/SoAy/tree/master/scenarios}.

\end{document}